\documentclass[11pt,a4paper]{article}
\usepackage[hyperref]{acl2017}
\usepackage{times}
\usepackage{latexsym}
\usepackage[british]{babel}

\usepackage{url}

\usepackage{CJKutf8}
\usepackage{float}
\usepackage{graphicx}
\usepackage{amsmath}
\usepackage{amssymb}
\usepackage{epsfig}
\usepackage{clrscode}
\usepackage{algorithm}
\usepackage{algpseudocode}
\usepackage{float}
\usepackage{multirow}
\usepackage{qtree}
\usepackage{epstopdf}
\usepackage{color}
\usepackage{mathrsfs}

\DeclareMathOperator*{\argmax}{argmax}

\usepackage{times}
\usepackage{comment}

\usepackage{tikz}
\usetikzlibrary{shapes,arrows, shadows}
\usetikzlibrary{calc,fit}

\aclfinalcopy 
\def\aclpaperid{118} 

\title{Tracing a Loose Wordhood for Chinese Input Method Engine}

\author{Xihu Zhang, Chu Wei\and Hai Zhao\thanks{*Corresponding author.}\\
  Department of Computer Science and Engineering,\\
  Shanghai Jiao Tong University, Shanghai, China \\
  {\tt \{xihuzhangcs, courage17340\}@gmail.com, zhaohai@cs.sjtu.edu.cn} \\}

\date{}

\begin{document}
\begin{CJK}{UTF8}{gbsn}
\maketitle
\begin{abstract}
  Chinese input methods are used to convert pinyin sequence or other Latin encoding systems into Chinese character sentences. For more effective pinyin-to-character conversion, typical Input Method Engines (IMEs) rely on a predefined vocabulary that demands manually maintenance on schedule. For the purpose of removing the inconvenient vocabulary setting, this work focuses on automatic wordhood acquisition by fully considering that Chinese inputting is a free human-computer interaction procedure. Instead of strictly defining words, a loose word likelihood is introduced for measuring how likely a character sequence can be a user-recognized word with respect to using IME. Then an online algorithm is proposed to adjust the word likelihood or generate new words by comparing user true choice for inputting and the algorithm prediction. The experimental results show that the proposed solution can agilely adapt to diverse typings and demonstrate performance approaching highly-optimized IME with fixed vocabulary.
\end{abstract}

\section{Introduction}

The inputting of Chinese language on computer differs from English and other alphabetic languages since Chinese language has more than 20,000 characters and they cannot be simply mapped to only 26 keys in the Latin keyboards. The Chinese Input Method Engines (IMEs) have been developed to aid Chinese inputting. Among all IMEs, the most prominent type is pinyin-based IME that converts pinyins, which are the phonetic representation of Chinese language, into Chinese character sequence. For example, a user wants to type a word "北京"(Beijing). The corresponding pinyin \emph{beijing} is inputted, then the pinyin IME provides a list of Chinese character candidates whose pinyins are all \emph{beijing}, such as
"北京"(pinyin: \emph{beijing}, the Beijing city), "背景"(pinyin: \emph{beijing}, background) as presented in Figure 1. The user at last selects the word "北京" as the result.

\begin{figure}[!htbp]
  \centering
\begin{tikzpicture}
\filldraw (0,0) node [below] {center here} circle (1pt);
\node [draw, rounded corners={0.05cm}, shape=rectangle, minimum width=5.0cm, minimum height=1.1cm, anchor=center,
            style={shade, top color=white, bottom color={rgb:blue,1;white,20}}] (what) at (0,0) {};
\node [below right] at ([shift={(0.03,-0.05)}]what.north west) {beijing};
\node [above right] at ([shift={(0.03,0.03)}]what.south west) {\small 1.北京 \ \ 2.背景 \ \ 3.背静 \ \ 4.被 \ \ 5.倍};
\end{tikzpicture}
  \caption{IME interface on one page}
\end{figure}

Chinese IMEs often have to utilize word-based language models since character-based Language Model does not produce satisfactory result~\cite{yang1998statistics}, Chinese words are composed of multiple consecutive Chinese characters. But unlike alphabetic languages, The words in Chinese sentences are not demarcated by spaces, making Chinese word identification a nontrivial task. Therefore, Chinese word dictionaries are usually predefined to support word-based language models for IMEs.

However, there is a huge difference between standard Chinese word segmentation and IME word identification. As Chinese text has no blanks between characters , it allows all possible consecutive characters to hopefully form Chinese words to some extent. For Chinese word segmentation task, it learns from segmented corpus or predefined segmentation conventions, therefore not all consecutive characters can be regarded as words for the task to learn\footnote{For example, a Chinese corpus may have 1 million bigrams, though still given fewer than 50 thousand meaningful word types.}. For IME word identification, it is a quite different case, because user seldom types a complete sentence and may stop at any position to leave a pinyin syllable segmentation and require IME to give proper pinyin-to-character conversion. Table 1 shows multiple possible segmentation choices for a given input. The segmentation points are indicated by vertical bar, though, the linguistically valid segmentation is only the first one,
which is "你\ \textbar\ 在\ \textbar\ 做\ \textbar\ 什么"(pinyin: \emph{ni zai zuo shenme}, what are you doing).

\begin{table}[!htbp]
\centering
\begin{tabular}{l}
{\color{black}1. \textbf{你(you)\ \textbar\ 在(be)\ \textbar\ 做(do)\ \textbar\ 什么(what)}}\\
{\color{black}2. 你(you)\ \textbar\ 在做(be doing)\ \textbar\ 什么(what)}\\
{\color{black}3. 你在(you are)\ \textbar\ 做什么(do what)}\\
{\color{black}4. 你(you)\ \textbar\ 在做什么(be doing what)}\\
{\color{black}5. 你在做什么(what are you doing)}\\
{\color{black}......}\\
\end{tabular}
\caption{IME user's input segmentations}
\end{table}

The converted consecutive characters forming `word' during inputting will not have a unified standard right as shown in Table 1, in which every segmentation in every items may be a `word' as any user-determined segment for IME inputting should be regarded as word. Thus word defined by user during Chinese inputting is actually a quite loose concept. Meanwhile, word from a linguistic sense is still a useful heuristic information for IME construction even though it is much more freely defined than that for the task of Chinese word segmentation. For example, IME being aware of that pinyin \emph{beijing} can be only converted to either "北京(Beijing)" or "背景(background)" will greatly help it make the right and more efficient pinyin-to-character decoding, as both pinyins \emph{bei} and \emph{jing} are respectively mapped to dozens of difference single Chinese characters (see Figure 2). Since capturing words can narrow down the scope of possible conversions, tracing such kind of wordhood for IME becomes important in improving user experiences.

\tikzset{
  treenode/.style = {draw=none},
  root/.style     = {treenode, font=\normalsize},
  env/.style      = {treenode, font=\ttfamily\small},
  dummy/.style    = {draw}
}
\begin{figure}[!htb]
\centering
\small
\begin{minipage}{.15\textwidth}
  \begin{tikzpicture}
  [
    grow                    = right,
    sibling distance        = 1.8em,
    level distance          = 2.5em,
    every node/.style       = {},
    sloped
  ]
  \node [root] {\emph{bei}}
    child { node [missing] (2) {}
      edge from parent[draw = none]}
    child { node [env] (1){\color{blue}背}
      edge from parent}
    child { node [env] {\color{blue}呗}
      edge from parent}
    child { node [env] {\color{red}\textbf{北}}
      edge from parent}
    child { node [env] {\color{blue}倍}
      edge from parent}
    child { node [env] {\color{blue}被}
      edge from parent};


    \path (1) -- node[auto=false,font=\large]{\dots} (2);
\end{tikzpicture}
 \end{minipage}
 \begin{minipage}{.15\textwidth}
  \begin{tikzpicture}
  [
    grow                    = right,
    sibling distance        = 2em,
    level distance          = 3em,
    every node/.style       = {font=\footnotesize},
    sloped
  ]
  \node [root] {\emph{jing}}
    child { node [missing] (2){}
      edge from parent[draw = none]}
    child { node [env] (1){\color{red}\textbf{京}}
      edge from parent}
    child { node [env] {\color{blue}竟}
      edge from parent}
    child { node [env] {\color{blue}精}
      edge from parent}
    child { node [env] {\color{blue}景}
      edge from parent}
    child { node [env] {\color{blue}静}
      edge from parent}
    child { node [env] {\color{blue}经}
      edge from parent};

    \path (1) -- node[auto=false,font=\large]{\dots} (2);

\end{tikzpicture}
 \end{minipage}
\begin{minipage}{.15\textwidth}
\begin{tikzpicture}
  [
    grow                    = right,
    sibling distance        = 4em,
    level distance          = 4em,
    every node/.style       = {font=\footnotesize},
    sloped
  ]
  \node [root] {\emph{beijing}}
    child { node [env] (1){\color{blue}背景}
      edge from parent}
    child { node [env] {\color{red}\textbf{北京}}
      edge from parent};

\end{tikzpicture}
\end{minipage}
\caption{All possible character counterparts of \emph{bei}, \emph{jing} and \emph{beijing}. Note that though modern Chinese has five tones, in all pinyin based IMEs, tones are not inputted with the letters for simplicity, therefore pinyin IME has to handle much more pinyin-to-character ambiguities in practice.}
\end{figure}
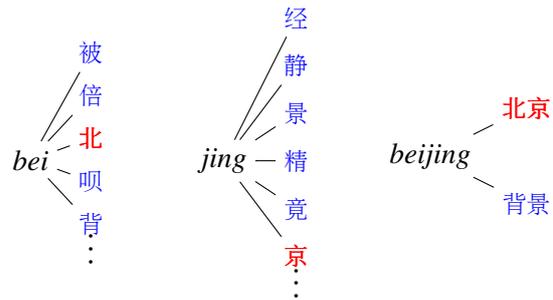

The above huge difference about wordhood definitions between Chinese word segmentation and IME results in most techniques developed for the former is unavailable to the latter, even we know that many studies have been paid to Chinese word segmentation, using either supervised, or unsupervised learning, and amazing results have been received~\cite{chang1997unsupervised,ge1999discovering,huang2003chinese,peng2002using,zhao2006effective,pei2014max,chen2015gated,ma2015accurate,cai2016segment}.

IMEs manage user inputting, which is a human-computer interaction software. We here consider two issues as follows for effective processing.

\begin{itemize}

\item \textbf{Challenge:} As a user behavior, Chinese input may vary from user to user and from time to time, which makes it impossible to find a stable and general model for the core components of IME, vocabulary and user preference. For example, a word may be frequently inputted by a user in a few days, but after then it is never touched any more.

\item \textbf{Convenience:} Every time, user makes a choice from character or character sequence candidate list given by IME, which makes inputting an online human-computer interaction procedure. It means that every time user indicates if the prediction given by IME is correct\footnote{IMEs are always capable of letting users input whatever they want, and the standard about goodness for IMEs is actually behind its speed, namely, the faster an IME lets user input, the better it is. Therefore the \textbf{correct} prediction means that IME ranks the user intended Chinese input at the first position of the candidate list. Ranking the expected input at top position means that user may input by default keystroke so that the inputting is the most efficient.}. However, as to our best knowledge, such kind of interactive property of IMEs has not been well studied or exploited. In fact, most, if not all, existing IMEs consider pinyin-to-character decoding in an offline model.

\end{itemize}

To solve the word dilemma in Chinese IME, this paper presents a novel solution of automatic wordhood acquisition for IMEs without assuming vocabulary initialization by fully considering that Chinese inputting is a human-computer interaction procedure for the first time. In detail, the proposed method uses an algorithm to trace word likelihood (but not word) that measures how a consecutive character sequence looks like a 'word' according to user inputting pattern. The algorithm initially only has a Pinyin-to-Character conversion table\footnote{There is a one-to-one mapping between a character and monosyllable pinyin for most Chinese characters, though a few Chinese characters may be mapped to two or more difference syllables. Such a mapping or conversion table is to define Chinese pinyin over difference Chinese characters.} and an empty vocabulary. Each time the algorithm receives a pinyin sequence and predicts its corresponding Chinese sequence. By comparing the user's true choice for inputting and the algorithm prediction, the vocabulary is updated and the newly-introduced word likelihood is adjusted to follow user inputting change.

Our paper is organized as follows. Section 2 displays related work. Section 3 introduces the framework of our model, discusses the pinyin-to-character conversion method and proposes an algorithm for adaptive word likelihood adjustment. In section 4 the experiment settings and results are demonstrated. Section 5 concludes the paper.

\section{Related Work}

Early work on IMEs include trigram statistical language model~\cite{chen2000new} and joint source-channel model~\cite{haizhou2004joint} that is applicable to Chinese, Japanese and Korean languages. Recent studies pay more attention on IME input typo correction ~\cite{zheng2011chime,jia2014joint} along with various approaches on Chinese spell checking~\cite{chen2013study,han2013maximum,chiu2013chinese}. There are also stuides on Janpanese IME which focus on kana-kanji conversion, including probabilistic language model based method~\cite{mori1998kana} , discriminative method~\cite{jiampojamarn2008joint,cherry2009discriminative,tokunaga2011discriminative}, improved $n$-pos model~\cite{chen2012using}, and ensemble method~\cite{okuno2012ensemble}. However, all the above are based on predefined vocabulary and did not exploit the interactive property of IME inputting. ~\cite{gao2002toward} and ~\cite{mori2006phoneme} tried to address unknown word problem, but they still require an initial vocabulary or a segmented training corpus.

In addition to the above studies over vocabulary organization and language model adaptation for IMEs, most current business IMEs actually use two engineering solutions to follow user vocabulary change. The first is to simply save and repeat what is frequently inputted and confirmed by user in recent times. Nevertheless, the recent high-frequency inputting will not be permanently added to the IME vocabulary. The second is to push new words or new language model on schedule. The new words are collected and mined from all online IME users' inputs. However, the updates about new words or new model are not real-time, and rely on a distributed user group, not adaptive for each individual user.

This work differs from all existing scientific and engineering solutions for the following aspects. First, we focus on adaptive word acquisition for IMEs, but not for any standard machine learning tasks such as all the recent work about Chinese word segmentation, as IME word definition is quite different from all these tasks~\cite{chang1997unsupervised,ge1999discovering,huang2003chinese,peng2002using,zhao2006effective,pei2014max,ma2015accurate,chen2015gated,cai2016segment}. Second, the interactive property of IMEs is to be fully exploited while was never seriously taken into account in previous work. Third, the proposed method will be optimized for each user but not all users' statistics and keep updates nearly at real time.

\section{Our Model}

The proposed model consists of two technical modules. First a Pinyin-to-Character Conversion Module receives pinyin string from user, and decodes it into a character sequence. Since IME provides conversion candidates, user can complete the correct character sequence with additional efforts. Therefore after user completes inputting, the IME gets correct character sequence corresponding to input pinyin string which may help improve further prediction.

\subsection{Pinyin-to-Character Conversion}

At the beginning, the input string is segmented into a pinyin syllable sequence. Chinese pinyin is highly regular, as every pinyin syllable contains a consonant and a vowel or just is a vowel-only syllable in a few simple types. The segmentation of pinyin syllables already has very high accuracy (over 98\%) by using a trigram language model for pinyin syllables with improved Kneser-Ney smoothing~\cite{kneser1995smoothing}\footnote{Pinyin language model can be conveniently trained on a pinyin syllable corpus.}
.

According to our previous discussion, word can be a very loose concept for IME in practice. Therefore, instead of defining an absolute wordhood, we introduce IME Word Likelihood (IWL) which can be regarded as a kind of wordhood weight for every consecutive character sequence to evaluate how possible it can be a word for IMEs.

For the conversion from pinyin syllables to character sequences, we are handling a kind of translation task from pinyin to character, therefore a translation model has to be built. Suppose that such a condition probability distribution has been learned, our objective is to find the character sequence $\hat{W}$ segmented by learned words which maximizes the conditional probability $\Pr[\hat{W}|P]$ given the input pinyin sequence $P$ with assuming that the sequence has been segmented into learned words and each word only depends on its previous $n-1$ word history.


\begin{figure*}[!htbp]
  \centering
  \includegraphics[width=14.0cm]{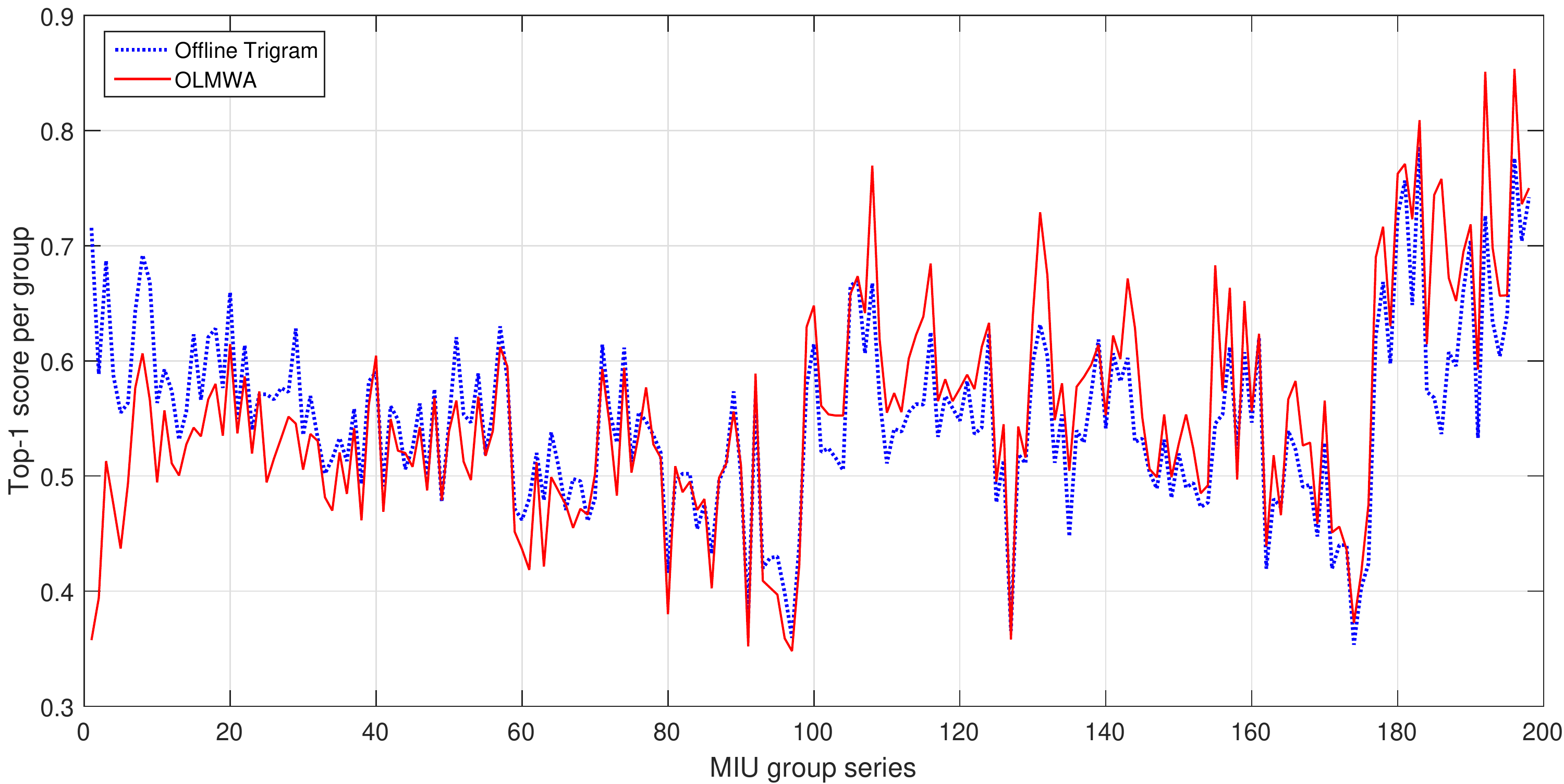}
  \caption{Top-$1$ score on \emph{the People's Daily}}
\end{figure*}

\begin{displaymath}
\begin{aligned}
\hat{W} & = \mathop{\argmax}_{W} \Pr[W|P] \\
        & = \mathop{\argmax}_{W}\frac{\Pr[P|W]Pr[W]}{\Pr[P]} \\
        & = \mathop{\argmax}_{W} \Pr[P|W]Pr[W] \\
        & = \mathop{\argmax}_{W=w_1..w_m} \prod_i\Pr[p_i|w_i]Pr[w_i|w_{i-1}..w_{i-n+1}]
\end{aligned}
\end{displaymath}
where all conditional probabilities can be estimated or derived by IWLs and word counts as follows,
\begin{displaymath}
\begin{aligned}
&\Pr[w_i] = \frac{\text{IWL}(w_i)}{\sum_{w}\text{IWL}(w)}\\
&\Pr[w_i|w_{i-1}..w_j] = \frac{Count(w_j..w_{i-1}w_i)}{\sum_{w}Count(w_j..w_{i-1}w)}\\
&\Pr[p_i|w_i] = \frac{\Pr[p_i,w_i]}{\Pr[w_i]}
\end{aligned}
\end{displaymath}

In the above formula, $\Pr[P]$ is a constant for IME decoding task during each turn of inputting so that the conditional probability can be equally computed by joint probability. Notice that $\Pr[p_i,w_i]$ can be derived from counting co-occurrence of $p_i$ and $w_i$.

As for all existing IMEs, $\Pr[w_i]$ is estimated by an offline $n$-gram language model with a fixed vocabulary trained from a corpus. Here we propose an online algorithm to estimated on a dynamic vocabulary through the IWL which will be introduced in the next subsection in detail.

\begin{algorithm}[!htb]
\caption{Online Model}
\textbf{Input: }
\begin{itemize}
\item Update times $N$;
\item Vocabulary $D$;
\item \emph{IWL} table;
\item Chinese character sequence $C$
\item Predefined vocabulary capacity $cap$;
\item Predefined vocabulary culling period $per$
\item Predefined maximum length of IME word $maxlen$;
\item Predefined update parameters $\alpha$,$\beta$,$\gamma$
\end{itemize}
\textbf{Output: }
\begin{itemize}
\item Updated \emph{IWL} table;
\item Updated vocabulary $D$
\end{itemize}
\begin{algorithmic}[1]
\State \If $N \mod per = 0$ \textbf{do}
\State \quad \While size of $D$ $>$ $cap$ \textbf{do}
\State \quad \quad $w$ = word with lowest IWL in $D$
\State \quad \quad \text{Remove $w$ from $D$ and set its IWL to 0}
\State \quad \textbf{end while}
\State \textbf{end if}
\State \For \text{all consecutive substring $s$ \textbf{in} $C$}
\State \quad \If length of $s \leq maxlen$ \textbf{do}
\State \quad \quad Add substring $s$ to $D$
\State \quad \quad $\emph{IWL}(s) \emph{+=} \alpha$
\State \quad \textbf{end if}
\State \textbf{end for}
\State Segment $C$ to $S=w_1...w_m$ with $\max\Pr[W]$
\State \For $w$ \textbf{in} \text{segmentation} $S$
\State \quad Increase \emph{IWL} of word $w$ by $\beta*\Pr[W]+\gamma$
\State Increase $n$-gram counts by 1 if occurring in $S$.
\\
\Return $\emph{IWL}$, $D$;
\end{algorithmic}
\end{algorithm}

\subsection{Tracing IME Word Likelihood}


A straightforward approach is to define the IWL as the frequency of character sequence. This strategy is intutive and effective but causes an issue. That is, the substring of words always has higher IWL even it is a meaningless fragment. For example, "总而言之(overall)" may have obtained a large IWL value. But its meaningless fragment "总而言" can have even higher IWL value. One way to address this problem is to do segmentation that maximizing language model probability $\Pr[W]$ over the current converted character sequence and set bonus on the IWL of words appearing in segmentation output. In such case, the segmentation operation does recognize the concerned part as a word. Since higher $\Pr[W]$ implies more reliability of segmentation, IWL is added by the $\Pr[W]$ multiplied by a constant weight $\beta$ to amplify its effect on IWL. In our implementation, we set $\alpha = 1.0$, $\beta = 5.0$ and $\gamma = 1.0$ according to our primary empirical results on related development datasets.

As the implementation of the proposed algorithm only has limited memory for use, we cull the vocabulary by removing IME words with too low IWL values periodically.

Our detailed algorithm is presented in Algorithm 1. It maintains a vocabulary which is empty initially. For every possible IME word in vocabulary, its IWL is updated online. For IME word which does not appear in vocabulary, its IWL is considered zero. The segmentation in line 13 that maximizes Pr[W] can be efficiently done by adopting Viterbi algorithm~\cite{viterbi1967error}.

\section{Experiments}

\subsection{Evaluation Metrics and Corpora}

\begin{table}[!htbp]
\centering
\small
\begin{tabular}{|lllllll|}
\hline
&zi&ran&yu&yan&chu&li\\
\hline
1.&自&然&语&言&处&理\\
2.&自&然&语&言&  &  \\
3.&自&然&  &  &  &  \\
4.&自&燃&  &  &  &  \\
5.&孜&然&  &  &  &  \\
\hline
\end{tabular}

\caption{Top-$5$ IME candiates for converting `Natural Language Processing'}
\end{table}

\begin{figure*}[!htbp]
  \centering
  \includegraphics[width=14.0cm]{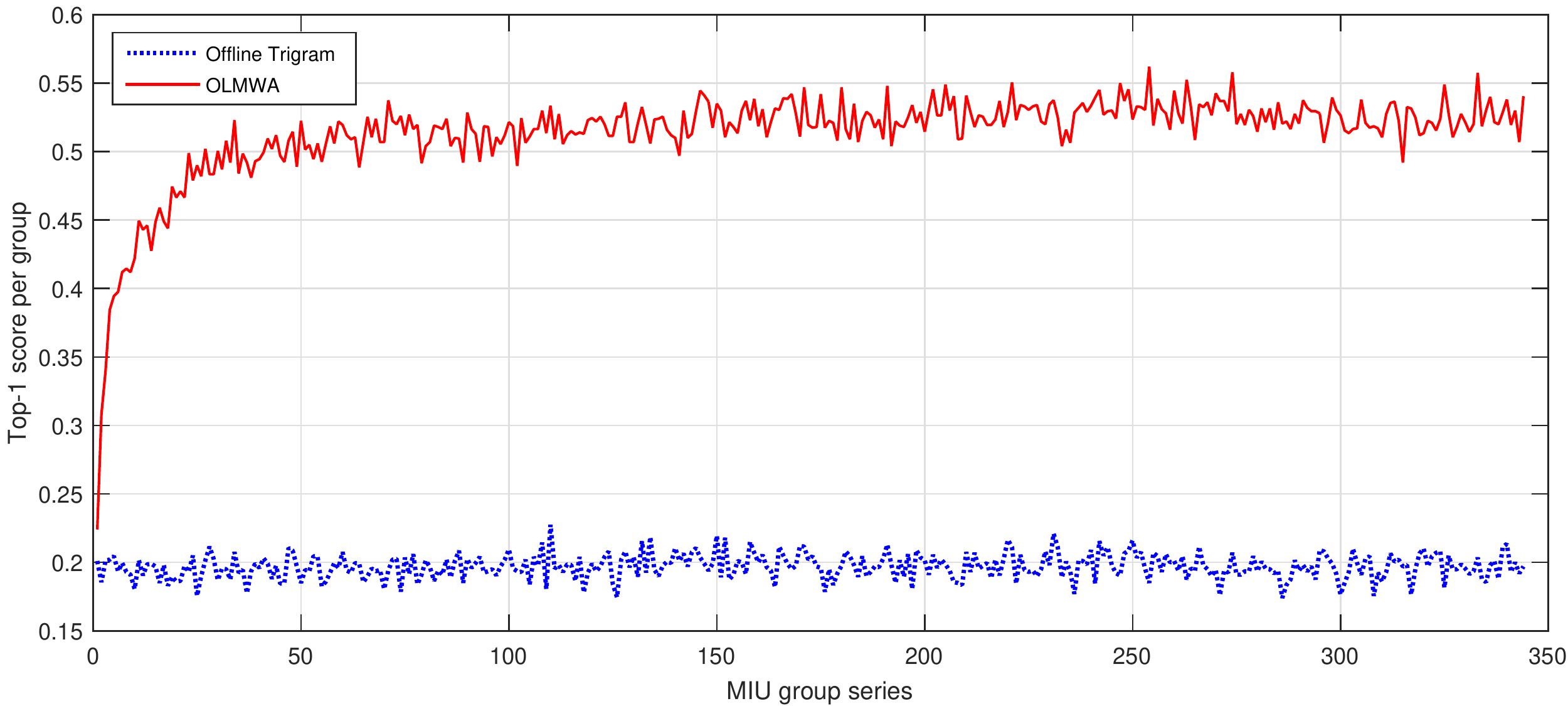}
  \caption{Top-$1$ score on \emph{Touchpal}}\label{fig:digit}
\end{figure*}

Chinese sentences may contain digits, alphabets and punctuations that can be inputted without IMEs. Since our concern is the accuracy of Pinyin-to-Chinese conversion, we evaluate the IME performance on Maximum Input Unit (MIU) proposed by Jia and Zhao~\shortcite{jia2013kyss}. MIU is defined as the longest consecutive Chinese character sequence in sentences that are separated by non-Chinese-character parts. Notice that in real world using, IMEs always give a list of character sequence candidates as in Table 2 for user choosing and candidates ranked behind will be shorter for more confident outputs. Therefore, measuring IME performance is equivalent to evaluating such a rank list.

In practice, more exact, longer candidates will be put more front in the list and less inputting effort taken by user indicates the IME performing better. Thus we use the following top-$K$ ranking score on top-$K$ predicted candidates ${\mathcal{P}_i}$, by introducing a decay factor, $1/2^{i-1}$.
\begin{displaymath}
\begin{aligned}
\mathcal{S} = \sum_{i=1}^{K}\frac{1}{2^{i-1}}I(\mathcal{P}_i, \mathcal{C})\frac{|\mathcal{P}_i|}{|\mathcal{C}|}
\end{aligned}
\end{displaymath}
where we use $|\mathcal{C}|$, $|\mathcal{P}_i|$ to denote the length of correct character sequence $\mathcal{C}$ and $\mathcal{P}_i$ respectively, and $I$ is a function as the following.
\begin{displaymath}
\begin{aligned}
I(\mathcal{P}_i, \mathcal{C}) =
\begin{cases}
1, &\mathcal{P}_i\text{ is prefix of }\mathcal{C} \\
0, & \text{otherwise}
\end{cases}
\end{aligned}
\end{displaymath}

In the following experiments we evaluate the performance on both top-$1$ and top-$10$ scores. The former is the most strict metric that directly measures the ratio of correct predicted MIUs that can be inputted by default (namely, the least) key strokes. The latter shows more complete candidates for choices and could better model the real IME using environment as IME basically gives a ranking list rather than a unique output. Note that IMEs can always give the expected inputting by user, the only difference between good and bad IMEs is about how they rank the input candidates for user to make the most convenient choice.


Two corpora presented in Table 3 is used for evaluation. The first corpus is from that previously used in~\cite{yang2012machine}, which has 390K MIUs for test set and 4M MIUs for training set. This corpus is extracted from \emph{the People's Daily}\footnote{This is the most popular newspaper in China.} from 1992 to 1998 that has word segmentation annotations by Peking University, also used for the second SIGHAN Bakeoff shared task~\cite{emerson2005second}. The pinyin annotations follow the method in~\cite{yang2012machine}. The second corpus is from Touchpal\footnote{Touchpal is a leading IME provider in the world-wide market, http://www.touchpal.com.} with 689K MIUs, which were collected from real user inputting history in recent years.

\subsection{Online vs. Offline}

\begin{figure*}[!htbp]
  \centering
  \includegraphics[width=14.0cm]{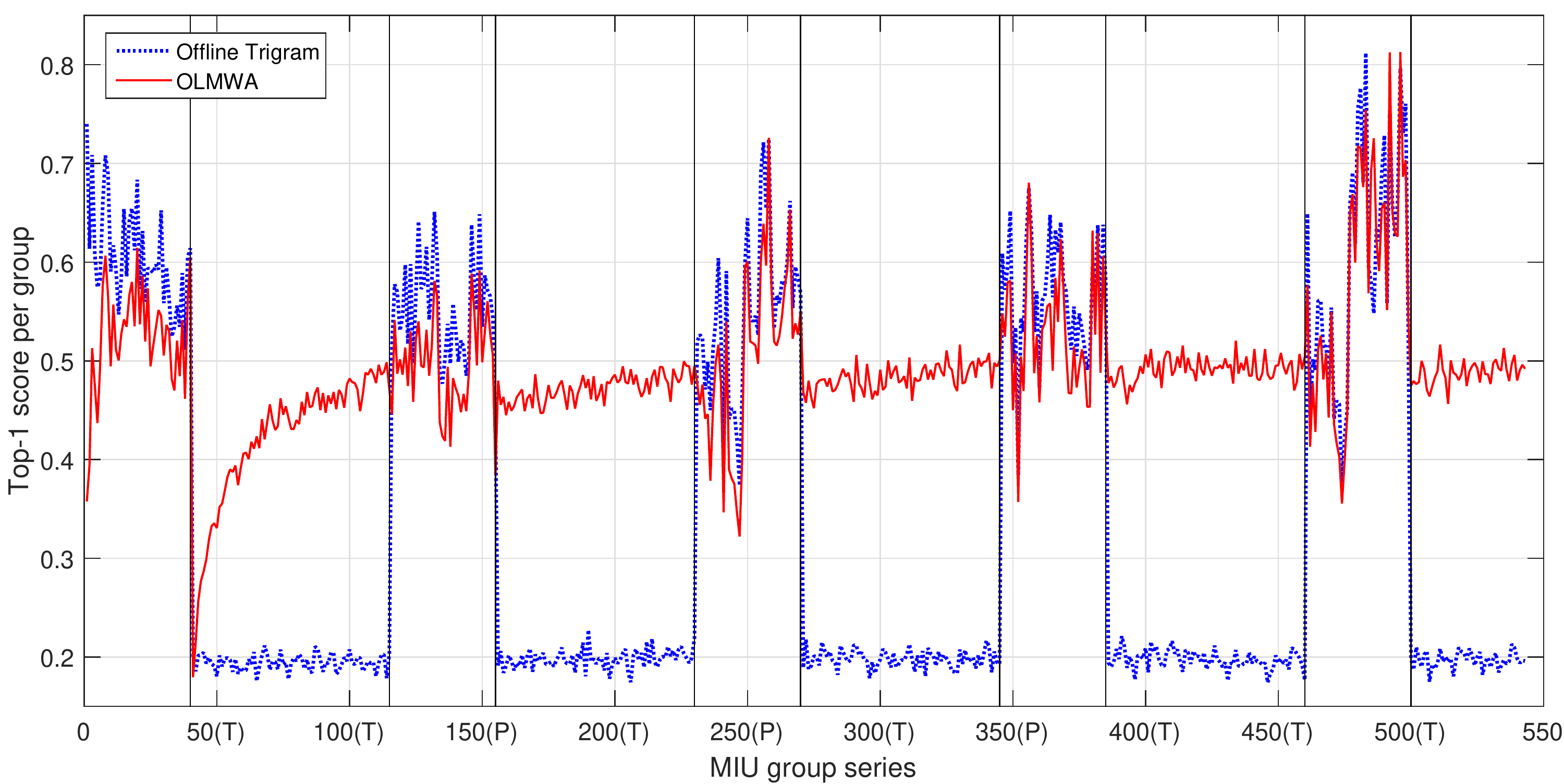}
  \caption{Top-$1$ score on \emph{Joint Corpus}. P: \emph{the People's daily segment}, T: \emph{Touchpal} segment}\label{fig:digit}
\end{figure*}

To show that our model can learn words automatically, our online model with empty vocabulary initialization is compared to the top-$K$ score with offline $n$-gram model which is without any smoothing techniques on large corpus.

Our Online Model for Word Acquisition (OMWA) is $4$-gram and initially assigned an empty vocabulary with $1$-million IME word capacity that stimulates the real case that only has limited memory. The OMWA is not trained on any corpus and we directly run it on the test set.

We will use test sets of \emph{the People's Daily} corpus and Touchpal corpus for evaluation. Note that these two corpora have a significant domain difference as the former is news about 20 years ago and the latter is about free inputs by various users in recent years.

\begin{table}[!htbp]
\centering
\begin{tabular}{p{2.8cm}|p{1.3cm}p{1.8cm}}
\hline
Corpus & \#MIUs & \#Characters\\
\hline
\emph{People's Daily} & \multicolumn{1}{r}{397,647} & \multicolumn{1}{r}{3,611,704}\\
\emph{Touchpal} & \multicolumn{1}{r}{689,692} & \multicolumn{1}{r}{5,820,749}\\
\hline
\end{tabular}
\caption{Corpus statistics}
\end{table}

We do comparison with three offline models respectively for unigram, bigram and trigram built on the training set of \emph{the People's Daily}. The vocabulary of offline models is directly extracted from the training set with segmentation annotation. The implementation of pinyin-to-character conversion with offline $n$-gram language model follows ~\cite{jia2014joint} without spelling correction for fair comparison.

The top-$K$ scores on \emph{the People's Daily} and \emph{Touchpal} are presented in Table 4, which demonstrates that OMWA outperforms all offline models even without vocabulary initialization. Here we show both top-$1$ and top-$10$ scores, in which the latter indicates two `pages' of ranking list for a typical IME output. On \emph{Touchpal}, OMWA demonstrates its learning capability and outperforms all offline models significantly. Note that all offline models and vocabulary are trained or extracted from \emph{the People's Daily} corpus, which indicates a domain difference. Offline models cannot work as well as it does on its in-domain corpus. In contrast, the proposed OMWA keeps outputting much more stable and better results than all offline models.

\begin{table}[!htbp]
\centering
\begin{tabular}{c|c|c|c|c}
\hline
Corpus & \multicolumn{2}{c|}{Method}                  & \multicolumn{1}{c}{Top1}   & Top10\\ \hline
\multirow{4}{*}{\begin{minipage}{0.5in}\emph{People's \\Daily}\end{minipage}}
                                & \multirow{3}{*}{Offline}    & Unigram      &  \multicolumn{1}{r}{35.55} & \multicolumn{1}{r}{51.88} \\
                                &                             & Bigram       & \multicolumn{1}{r}{51.78} & \multicolumn{1}{r}{69.00} \\
                                &                             & Trigram      & \multicolumn{1}{r}{54.96} & \multicolumn{1}{r}{70.14} \\ \cline{2-3}
                                & \multicolumn{2}{c|}{OMWA}  & \multicolumn{1}{r}{\textbf{55.27}} & \multicolumn{1}{r}{\textbf{74.25}}\\
\hline
\multirow{4}{*}{\emph{Touchpal}}
                                & \multirow{3}{*}{Offline}    & Unigram      &  \multicolumn{1}{r}{8.30} & \multicolumn{1}{r}{19.77} \\
                                &                             & Bigram       & \multicolumn{1}{r}{17.68} & \multicolumn{1}{r}{26.92} \\
                                &                             & Trigram      & \multicolumn{1}{r}{19.70} & \multicolumn{1}{r}{27.73} \\ \cline{2-3}
                                & \multicolumn{2}{c|}{OMWA}  & \multicolumn{1}{r}{\textbf{51.40}} & \multicolumn{1}{r}{\textbf{78.94}}\\
\hline
\end{tabular}
\caption{Comparison with offline models trained on \emph{the People's Daily} only (\%)}\label{my-label}
\end{table}

As the proposed OMWA relies on the interactive property of inputting, we also measure top-$1$ score over each 2K sentences to demonstrate how OMWA works over these MIU group series. Since trigram generally outperforms unigram and bigram, we only present the comparison between OMWA and offline trigram. The top-$1$ score per group on \emph{the People's Daily} is presented in Figure 3 and the top-$1$ score per group on \emph{Touchpal} is presented in Figure 4.

Figure 3 shows how OMWA goes after the offline trigram model which is trained right at the in-domain corpus. After learning from a few sentences, OMWA quickly approaches the offline trigram. After half a corpus has been learned, OMWA generally outperforms the offline trigram model.

Figure 4 shows that OMWA keeps outperforming offline trigram model. One of the reasons is the trigram model is an out-of-domain model for \emph{Touchpal} corpus as it is trained on \emph{the People's Daily} and it cannot adapt a different corpus.

To demonstrate the robustness of OMWA, we present top-$1$ score per group on a combined corpus in Figure 5. Both corpora, \emph{the People's daily} and \emph{Touchpal}, are divided into five segments, then they are interlaced into one corpus. Vertical lines in the figure indicate joints between two different corpora. From every joints including the very begining, OMWA may adapt the corpus change after several sentences are learned. As the learning accumulation has been more and more, the adaptation of OMWA becomes more and more quick for each change. On the contrary, the offline trigram model performs stably only on its in-domain segments.
Overall, OMWA has demonstrated robust performance for all corpus changes.

\subsection{Comparison with Google IME}

We run the OMWA on corpus training set and use the final model to compare with \emph{Google Input Tools}\footnote{http://www.google.com/inputtools/try} since it is the only commercial IME that provides accessible interface for us to do comparison\footnote{We are aware that there are many more popular commercial IMEs. However, Google IME is currently the only choice that provides an applicable interface to enable us to make a fair comparison.}. Our model is only a simple model for word likelihood tracing while Google IME is a commercial model which has utilized various components including optimized language model, high-quality vocabulary and massive corpora. Due to the connection limitation of Internet, we only run testing on a small part of corpus. The corpora used in this experiment are shown in the Table 5,and the results are in Table 6.

\begin{table}[!htbp]
\centering
\begin{tabular}{l|l|ll}
\hline
\multicolumn{2}{l|}{}                  & People's Daily          & TouchPal          \\ \hline
\multirow{2}{*}{Training}         & \#MIUs         &  \multicolumn{1}{r}{4,231,352} & \multicolumn{1}{r}{689,692} \\
                           & \#Chars         & \multicolumn{1}{r}{38,236,958} & \multicolumn{1}{r}{5,820,749} \\ \hline
\multirow{2}{*}{Testing}         & \#MIUs         & \multicolumn{1}{r}{7,603} & \multicolumn{1}{r}{6,998} \\
                           & \#Chars         & \multicolumn{1}{r}{74,277} & \multicolumn{1}{r}{59,151} \\
\hline
\end{tabular}
\caption{Corpora for Google IME comparison}\label{my-label}
\end{table}

\begin{table}[!htbp]
\centering
\begin{tabular}{c|c|c|c}
\hline
Corpus & Method                  & \multicolumn{1}{c}{Top1}   & Top10\\ \hline
\multirow{2}{*}{\begin{minipage}{0.5in}\emph{People's \\Daily}\end{minipage}}
                                & Google      & \multicolumn{1}{r}{\textbf{70.93}} & \multicolumn{1}{r}{\textbf{82.25}} \\ \cline{2-2}
                                & OMWA       & \multicolumn{1}{r}{64.41} & \multicolumn{1}{r}{77.93}\\
\hline
\multirow{2}{*}{\emph{Touchpal}}
                                & Google      & \multicolumn{1}{r}{\textbf{57.50}} & \multicolumn{1}{r}{69.34} \\ \cline{2-2}
                                & OMWA       & \multicolumn{1}{r}{57.12} & \multicolumn{1}{r}{\textbf{80.95}}\\
\hline
\end{tabular}
\caption{TopK score comparison with Google IME(\%)}\label{my-label}
\end{table}

The comparison shows that OMWA has achieved competitive top-$1$ score and better top-$10$ score comparing to Google IME on \emph{Touchpal}, but relative lower performance on \emph{People's daily}. The dramatic difference of performance between two corpora by Google IME may indicate an interesting discovery. As well known, the \emph{People's Daily} corpus has been popularly used in computational linguistics research and industry. It is no doubt that Google IME was made especial optimization on the popular corpus, but failed to do so on another private corpus in a different domain. Being only a simple learner, OMWA performs similarly as Google IME on a corpus that the latter is not familiar with, which has demonstrated the proposed model alone can be comparable to a commercial IME and it still keeps the potential to be better through later much more learning.

\subsection{IME Words vs. Linguistic Words}

Finally we present words extracted by OMWA with top word likelihood weights from \emph{the People's Daily} in Table 7. It is interesting that these top weighted words discovered by OMWA are indeed true words according to strict Chinese linguistic standard.

\begin{table}[!htbp]
\centering
\begin{tabular}{p{4.2cm}|p{1.6cm}}
\hline
Word & IWL\\
\hline
中国(China) & 13791 \\
发展(develop) & 11430\\
工作(job) & 9805 \\
经济(economy) & 9579 \\
建设(construct) & 8139 \\
人民(people) & 7761 \\
同志(comrade) & 7657 \\
社会(society) & 7572 \\
企业(enterprise) & 7471 \\
国家(country) & 6934 \\
\hline
\end{tabular}
\caption{Words learned with top word likelihood weights on \emph{the People's Daily}}
\end{table}

\section{Conclusion and Further Discussions}
In this paper, we proposed an Online Model for Word Acquisition. This algorithm adaptively learns new words, or more precisely, new word likelihood, for more efficient inputting as using Chinese IME. The experiments show that the proposed method can achieve competitive result compared to commericial IME, and demonstrates more robustness and more adaptivity for changes of user inputting style.

In addition, the proposed method is more than an online word learner only for IME, it shows standard language model can be flexible enough to defined on a slightly different ways, in which all $n$-grams can equally receive probability updating, without considering they are either known or unknown, so that the strict constraint about fixed vocabulary for language training can be released.

At last, the proposed learner with a linear updating strategy can be further improved by adopting neural popular neural models for potential performance enhancement, which will be left for the future work.

\bibliography{acl2017}
\bibliographystyle{acl_natbib}

\end{CJK}
\end{document}